\documentclass{IEEEtran}
\usepackage{cite}
\usepackage{amsmath,amssymb,amsfonts}
\usepackage{algorithmic}
\usepackage{graphicx}
\usepackage{textcomp}

\usepackage{graphics}
\usepackage{subcaption}
\usepackage{amsthm,amsmath}
\usepackage{authblk}
\usepackage{multirow}
\usepackage[]{algorithm2e}
\usepackage[utf8]{inputenc} 
\usepackage[T1]{fontenc}    

\usepackage{url}

\begin{document}

\title{Cleora: A Simple, Strong and Scalable Graph Embedding Scheme}

\author[12]{Barbara Rychalska}
\author[1]{Piotr Bąbel}
\author[1]{Konrad Gołuchowski}
\author[1]{Andrzej Michałowski}
\author[1]{Jacek Dąbrowski}
\affil[1]{Synerise}
\affil[2]{Warsaw University of Technology}

\maketitle

\begin{abstract}
The area of graph embeddings is currently dominated by contrastive learning methods, which demand formulation of an explicit objective function and sampling of positive and negative examples. This creates a conceptual and computational overhead. Simple, classic unsupervised approaches like Multidimensional Scaling (MSD) or the Laplacian eigenmap skip the necessity of tedious objective optimization, directly exploiting data geometry. Unfortunately, their reliance on very costly operations such as matrix eigendecomposition make them unable to scale to large graphs that are common in today's digital world. In this paper we present Cleora: an algorithm which gets the best of two worlds, being both unsupervised and highly scalable. We show that high quality embeddings can be produced without the popular step-wise learning framework with example sampling. An intuitive learning objective of our algorithm is that a node should be similar to its neighbors, without explicitly pushing disconnected nodes apart. The objective is achieved by iterative weighted averaging of node neigbors' embeddings, followed by normalization across dimensions. Thanks to the averaging operation the algorithm makes rapid strides across the embedding space and usually reaches optimal embeddings in just a few iterations. Cleora runs faster than other state-of-the-art CPU algorithms and produces embeddings of competitive quality as measured on downstream tasks: link prediction and node classification. We show that Cleora learns a data abstraction that is similar to contrastive methods, yet at much lower computational cost. We open-source Cleora under the MIT license allowing commercial use under  \underline{\url{https://github.com/Synerise/cleora}}.
\end{abstract}

\section{Introduction}
\label{sec:introduction}
 Graphs are data structures which are extremely useful for modeling real-life interaction structures. A graph is represented by sets of nodes and edges, where each node represents an entity from the graph domain, and each edge represents the relationship between two or more nodes. Graph structures are found for example in biology as interconnected sets of amino acids building proteins  \cite{10.1093/bioinformatics/btz600, yue2020graph, ingraham2019generative, yao2019accurately}, road networks \cite{wu2020learning, zheng2020gman}, as well as social networks \cite{aletras2018predicting, zhang2018cosine}, citation networks \cite{pornprasit2020convcn, asatani2018detecting}, or web data \cite{ristoski2016rdf2vec, cochez2017biased}. In most machine learning applications it is crucial to represent graph nodes as node embeddings - structures expressing node properties as an input to downstream machine learning algorithms. A simple node adjacency matrix is usually not feasible due to its large size (quadratic with respect to the number of nodes) and lack of easily accessible representation of node properties. A number of elaborate node embedding methods have been proposed, with configurable embedding dimensionality and node similarity defined in terms of a selected distance metric. However, most of these approaches do not scale to the real-world large graphs consisting of millions of nodes and billions of edges. Methods such as Deepwalk \cite{perozzi2014deepwalk} or Node2Vec \cite{grover2016node2vec}, require hours or even days of training even on medium-sized graphs to arrive at a representation of reasonable quality. Scalable models have also been proposed, such as PyTorch BigGraph (PBG) \cite{pbg} or GOSH \cite{10.1145/3404397.3404456}, however such optimization methods demand significant model complexity to parallelize the computations or/and coarsen the graph into a smaller structure.

In this paper we present Cleora: a very simple yet competitive graph embedding scheme. Cleora relies on multiple iterations of normalized, weighted averaging of each node's neighbor embeddings. Each embedding dimension is optimized independently from other dimensions (with the exception of normalization, which takes into account all dimensions). As such, Cleora can be thought of as an ensemble of 1-D optimizers rather than a single optimizer of N-D data. The operation of averaging can make rapid, substantial changes to embeddings (as compared to step-wise optimization based on an objective), making the algorithm reach optimal embeddings in just four to five iterations on average. Cleora is purely unsupervised. No explicit learning objective is formulated and no sampling of positive or negative examples is performed.

Cleora has only two configurable parameters: the number of iterations and dimensionality of resulting embeddings, which makes it easy to search for an optimal configuration. Contrastive learning methods will usually have many more configurable parameters, for example PBG has at least 18 parameters which directly define the training process. Moreover, some methods (PBG included) define a number of loss functions which optimize embeddings for various purposes separately (e.g. node ranking or classification). With Cleora the obtained embeddings are versatile as no explicit objective is defined.

Cleora scales well to massive graph sizes and can embed various types of edges: undirected, directed and/or weighted. We evaluate Cleora on various real-world datasets of medium to large size, in order to show that the quality of  produced  embeddings is competitive to recent state-of-the-art methods, without the overhead of complex architecture. We show that Cleora is faster than other recent CPU-based methods, thanks to the simplicity and the ease of parallelization.

Moreover, our algorithm possesses two interesting properties which are a consequence of the weighted averaging approach and may turn up useful in practical scenarios:
\begin{itemize}
    \item \textbf{Additivity.} The embedding procedure can be conducted on chunked graph and the embedded chunks can be merged with a single Cleora step, without any extra solution. As the method is based on simple weighted averaging, an average of all chunks' embeddings produces the final embeddings. This simply repeats the operation which would have happened on the full graph if embedded directly. Cleora can embed massive graphs which do not fit into RAM/disk space at once.
    
    \item \textbf{Inductivity.} Embedding of nodes added after the computation of full graph embedding is a natural operation, requiring a single iteration of the algorithm on just the newly added nodes. None of the competitors we evaluate have this ability.
    
    
\end{itemize}

Cleora is an algorithm which processes data from large retailers in our production environment, with graphs comprised of millions of nodes and billions of edges. Embedding times for our biggest e-commerce datasets are below 2 hours on a single Azure virtual machine of type Standard E32s v3 32 vCPUs/16 cores and 256 GB RAM. Cleora has already been in use for 1 year.

We release Cleora as open-source software. We offer an easy to run, highly optimized implementation written in Rust. We release the code under the permissive MIT license, which allows commercial use\footnote{\url{https://github.com/Synerise/cleora}}.

\section{Related Work}
\label{related-work}

The first attempts at graph embeddings were purely unsupervised. Methods such as classic PCA \cite{Jolliffe1986}, Laplacian eigenmap \cite{NIPS2001_1961}, MSD \cite{cox2000multidimensional}, and IsoMap \cite{tenenbaum_global_2000} typically exploit spectral properties of various matrix representations of graphs, such as the Laplacian and the adjacency matrix. In essence, these methods preform dimensionality reduction while preserving distance relations between nodes (viewed as path distances in graphs). Unfortunately, their complexity is at least quadratic to the number of nodes, which makes these methods far too slow for today's volumes of graph data. 

A subsequent, essential line of work was based on random walks between graph nodes. The classic DeepWalk model \cite{perozzi2014deepwalk} uses random walks to create a set of node sequences fed to a skip-gram model, taking an inspiration from learning representations of tokens in Word2Vec \cite{Word2Vec} from the area of natural language processing. Node2vec \cite{grover2016node2vec} is a variation of DeepWalk where the level of random walk exploration (depth-first search) versus exploitation (breadth-first search) is controlled with parameters. Other more recent approaches make the random walk strategy more flexible at the cost of increased complexity \cite{harp, 10.1145/3110025.3110086}. These models perform well in practice, but cannot scale to graphs with millions of nodes.  

LINE \cite{10.1145/2736277.2741093} is yet another approach, which aims to preserve the first-order or second-order proximity separately by the use of KL-divergence minimization.  After optimizing the loss functions, it concatenates both representations. LINE is optimized to handle large graphs, but its GPU-based  implementation - Graphvite \cite{Zhu2019GraphViteAH} - cannot embed a graph when the total size of the embedding is larger than the total available GPU memory. With GPU memory size usually much smaller than available RAM, this is a serious limitation.

Another established line of work focuses on embedding graphs with the use of graph  convolutional neural networks \cite{Kipf:2016tc}. Methods such as 
GraphSAGE \cite{NIPS2017_6703} are designed for cases where the graph nodes are accompanied by additional feature information and incorporation of no-feature nodes might be complicated or demand additional measures. In contrast, our approach focuses on pure graph structure. 

Works closest to our aims in terms of scalability and speed are able to embed massive graphs in reasonable time without the requirement of node features or any kind of side information. These fast embedding methods often boil down to older methods implemented in parallelizable and highly efficient architectures. Pytorch BigGraph (PBG) \cite{pbg} uses ideas such as adjacency matrix partitioning and reusing examples within one batch to train models analogous to successful but less scalable RESCAL \cite{10.5555/3104482.3104584}, DistMult \cite{yang2014embedding}, TransE \cite{bordes2013translating},and ComplEx \cite{trouillon2016complex}, thus making these models applicable to large graphs. GOSH \cite{10.1145/3404397.3404456} is a GPU-based approach which trains a parallelized version of Verse \cite{verse} on a number of smaller, coarsened graphs. Embeddings are computed with Noise Contrastive Estimation between positive and negative samples, first on smaller, coarsened graphs and projected to bigger ones. In contrast to these methods, Cleora is not a reimplementation of an older approach but a conceptually new algorithm. Moreover, the listed methods are complex, demand step-wise training via gradient descent, and do not naturally generalize to unseen data.

\section{Cleora Embeddings}

\subsection{Preliminaries} A graph $G$ is a pair $(V, E)$ where $V$ denotes a set of nodes (also called vertices) and $E \subseteq (V \times V)$ as the set of edges connecting the nodes. We consider undirected graphs, where an edge is an unordered pair of nodes. An embedding of a graph $G=(V,E)$ is a $|V| \times d$ matrix $T$, where $d$ is the dimension of the embedding. The i-th row of matrix $T$ (denoted as $T_{i,*}$) corresponds to a node $i \in V$  and each value $j \in \{1, ..., d\}$ in the vector $T_{i,j}$ captures a different feature of node $i$.

Hypergraphs are a generalization of graphs where the relationships between nodes can be not only pairwise, but also higher-order. For example, the relation of Customer $c$ buying a Product $p$ in Store $s$ is a hyperedge composed of nodes ${c, p, s}$. Hyperedge width $k$ is defined as the number of nodes contained in the hyperedge. For example, the hyperedge $(c, p, s)$ has width 3.

\subsection{The Algorithm}

\textbf{Hypergraph Expansion.} 
  In order to produce hypergraph embedding, Cleora needs to break down all existing hyperedges into edges as the algorithm relies on the pairwise notion of node transition. Hypergraph expansion to graph is done using two alternative strategies:
\begin{itemize}
    \item \textit{Clique Expansion.} Each hyperedge is transformed into a clique - a subgraph where each pair of nodes is connected with an edge. Space/time complexity of the whole embedding procedure in this approach is given by $O(|V| \times d + |E| \times k^2)$ where $|E|$ is the number of hyperedges and $k$ is the maximal width of hyperedge from the hypergraph. With the usage of cliques the number of created edges can be significant but guarantees better fidelity to the original hyperedge relationship. We apply this scheme to smaller graphs.
    \item \textit{Star Expansion.} An extra node is introduced which links to the original nodes contained by a hyperedge. Space/time complexity of the whole embedding procedure in this approach is given by $O((|V|+|E|) \times d + |E|k)$. Here we must include the time and space needed to embed an extra entity for each hyperedge, but we save on the number of created edges, which would be only $k$ for each hyperedge. This approach is recommended for large graphs or graphs with large hyperedges.
\end{itemize}

\textbf{Embedding.} With all hyperedges broken down into pairwise edges, we proceed to embed the graph. Given an interaction network (e.g. between users and items) represented as a graph
$G=(V,E)$, we define the random walk transition matrix $\mathbf{M}\in \mathbb{R}^{V \times V} $ where $\mathbf{M}_{a b}=\frac{e_{a b}}{deg_(v_{a})}$ for $a b \in E$, where $e_{ab}$ is the
number of edges running from node $a$ to $b$, and $deg(v_{a})$ is the degree of node $a$. For $a b$ pairs which do not exist in the graph, we set $e_{a b} = 0$.
If edges have attached weights, $e_{ab}$ is the sum of the weights of all participating edges. 

We initialize the embedding matrix $\mathbf{T}_0 \in \mathbb{R}^{|V| \times d} \sim U\left(-1, 1\right)$, where $d$ is the embedding dimensionality. Then for $I$ iterations, we multiply matrix $\mathbf{M}$ and $T_i$, and normalize rows to keep constant $L_2$ norm. $\mathbf{T}_I$ is our final embedding matrix. If the graph is too big to be embedded at once due to memory constraints, it can be split and merged in a natural way by weighted averaging of the composing parts' embeddings. The full procedure (including graph splitting and averaging) is show in detail in Algorithm \ref{algo-cleora}.

\begin{algorithm}[]
 \KwData{Graph $G = (V,E)$ with set of nodes $V$ and set of edges $E$, iteration number $I$, chunk number $Q$, embedding dimensionality $d$} 
 \KwResult{Embedding matrix  $\mathbf{T} \in \mathbb{R}^{|V| \times d}$ }
Divide graph $G$ into $Q$ chunks. Let $G_{q}$ be the $q$-th chunk with edge set $E_{q}$\;

\For{q from 1 to Q} {
 For graph chunk $G_{q}$ compute random walk transition matrix $\mathbf{M}_{q}=\frac{e_{a b}}{deg(v_{a})}$ for $a b \in E_{q}$, where $e_{ab}$ is the number of edges running from node $a$ to $b$  \;
 Initialize chunk embedding matrix $\mathbf{T}^{q}_{0} \sim U\left(-1, 1\right)$ \;
 \For{i from 1 to I}{
    $\mathbf{T}^{q}_{i}=\mathbf{M}_{q} \cdot \mathbf{T}^{q}_{i-1}$\;
    $L_2$ normalize rows of $\mathbf{T}^{q}_{i}$\;
 }
 $\mathbf{T}^{q}$ = $\mathbf{T}^{q}_{I}$\;
 }
 
 \For{v from 1 to $|V|$} {
    $\mathbf{T_{v,*}}$ = $\sum_{q=1}^{Q}{w_{qv} \times \mathbf{T^{q}_{v,*}}}$ where $w_{qv}$ is the node weight $w_{qv} = \frac{|n \in V_{q}: n = n_{v} |}{\sum_{k=1}^{Q}{|n \in V_{k}: n = n_{v} |}}$
 }
 \caption{Cleora algorithm.}
 \label{algo-cleora}
\end{algorithm}

Initial vectors in matrix $\mathbf{T}_0$ need to 1) be different from each other so that the algorithm does not collapse to the same representation 2) have similar pairwise distances in order to avoid 'false friends' which are accidentally close in the embedding space. Matrix initialization from the uniform distribution creates vectors which fulfill these conditions in high dimensional space.

Cleora works analogously to an ensemble of $d$ models, as each dimension is optimized separately from others (see the column-wise matrix multiplication operation in Figure \ref{fig_implementation}. The only operation which takes into account all dimensions (and as such, allows some information sharing) is the $L_2$ normalization. The normalization step ensures numeric stability, preventing the norm of embedding vectors from collapsing towards zero during repeated multiplication by the transition matrix, of which the determinant is $\leq{1}$.

The method can be interpreted as iterated $L_2$-normalized weighted averaging of neighboring nodes' representations. After just one iteration, nodes with similar 1-hop neighborhoods in the network will have similar representations. Further iterations extend the same principle to q-hop neighborhoods. 

Thanks to the weighted averaging approach two useful operations become natural: 
\begin{itemize}
    \item Creation of new node embedding by weighted averaging of all neighbors' representations. With $I$ being the maximum iteration number, the representations should ideally come from $(I-1)$th iteration to mirror the last matrix multiplication and average. Using representations from $I$th iteration is not an error either, but caution is advised as the embedding quality can deteriorate quickly with increasing iteration number (see Section x).
    \item Obtaining node embedding from a large partitioned graph by weighted averaging of the representation from each slice.
\end{itemize}

\section{Implementation}

Cleora is implemented in Rust in a highly parallel architecture, using multithreading and adequate data arrangement for fast CPU access. 

\subsection{Modes of Operation}

Cleora ingests text files with relational tables of rows representing a heterogeneous hypergraph. The same input will be treated differently depending on parameters, which define the meaning of particular columns, and as such - the actual graph that will be created. Each column of nodes in the input file can be defined with a number of properties represented with column modifiers. Following column modifiers are available:
\begin{itemize}
    \item \texttt{transient} - the field is virtual - it is considered during embedding process, no entity is written for the column
    \item \texttt{complex} - the field is composite, containing multiple entity identifiers separated by space in TSV or an array in JSON
    \item \texttt{reflexive} - the field is reflexive, which means that it interacts with itself, additional output file is written for every such field
    \item \texttt{ignore} - the field is ignored, no output file is written for the field
\end{itemize}

\begin{figure}%
\centering
\includegraphics[width=80mm]{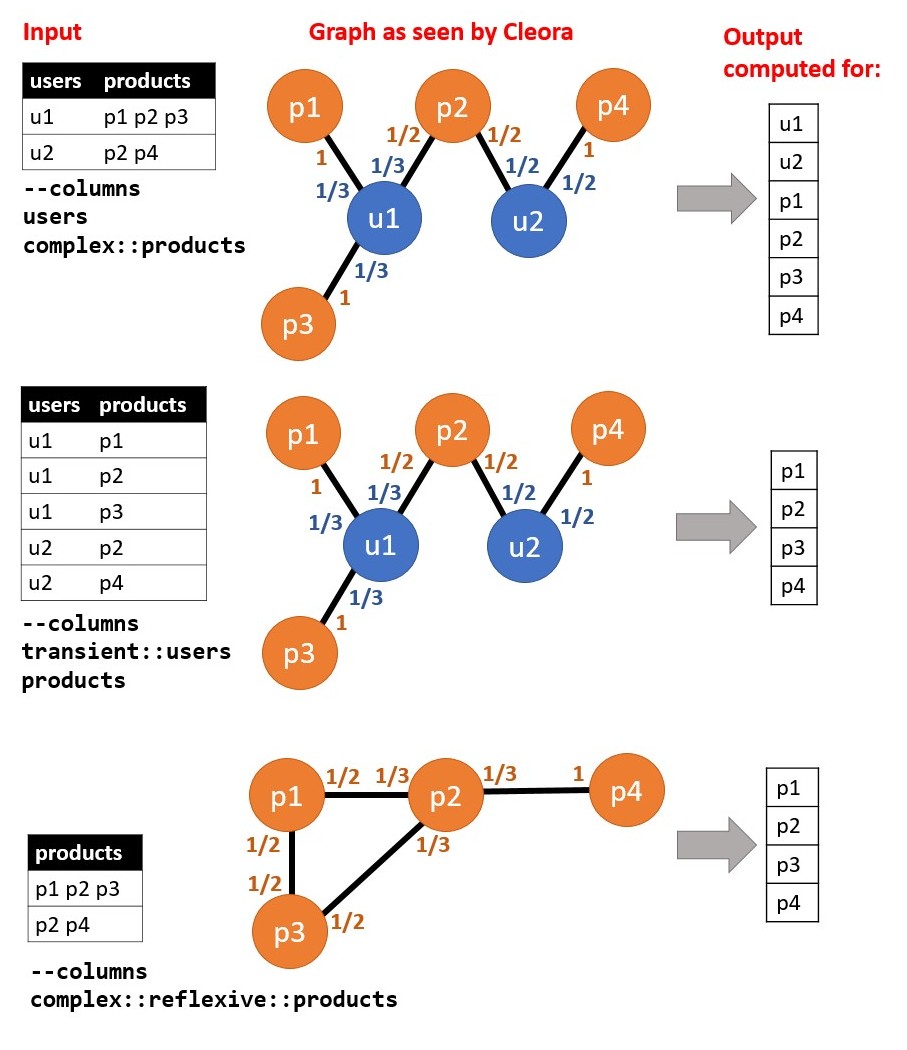}
\caption{Examples of various embedding modes and the resulting graph structures and output embeddings. Fractional numbers given next to nodes show values form transition matrix \textbf{M} for each node.}
\label{fig_cleora-example}
\end{figure}

Figure \ref{fig_cleora-example} shows examples of the most common combinations of column modifiers, with resulting graph structures and the outputs which will be written.

\subsection{Embedding Computation}
We exemplify the embedding procedure in Figure \ref{fig_implementation}, using a very general example of multiple relations in one graph. Embedding is done in two basic steps: graph construction and training. Such multi-relation configuration is allowed and will result in computation of a separate embedding matrix for each relation pair.

For maximum efficiency we created a custom implementation of a sparse matrix data structure - the \texttt{SparseMatrix} struct. It follows the sparse matrix coordinate list format (COO). Its purpose is to save space by holding only the coordinates and values of nonzero entities.

\subsection{Graph construction} 
\begin{figure*}%
\centering
\includegraphics[width=180mm]{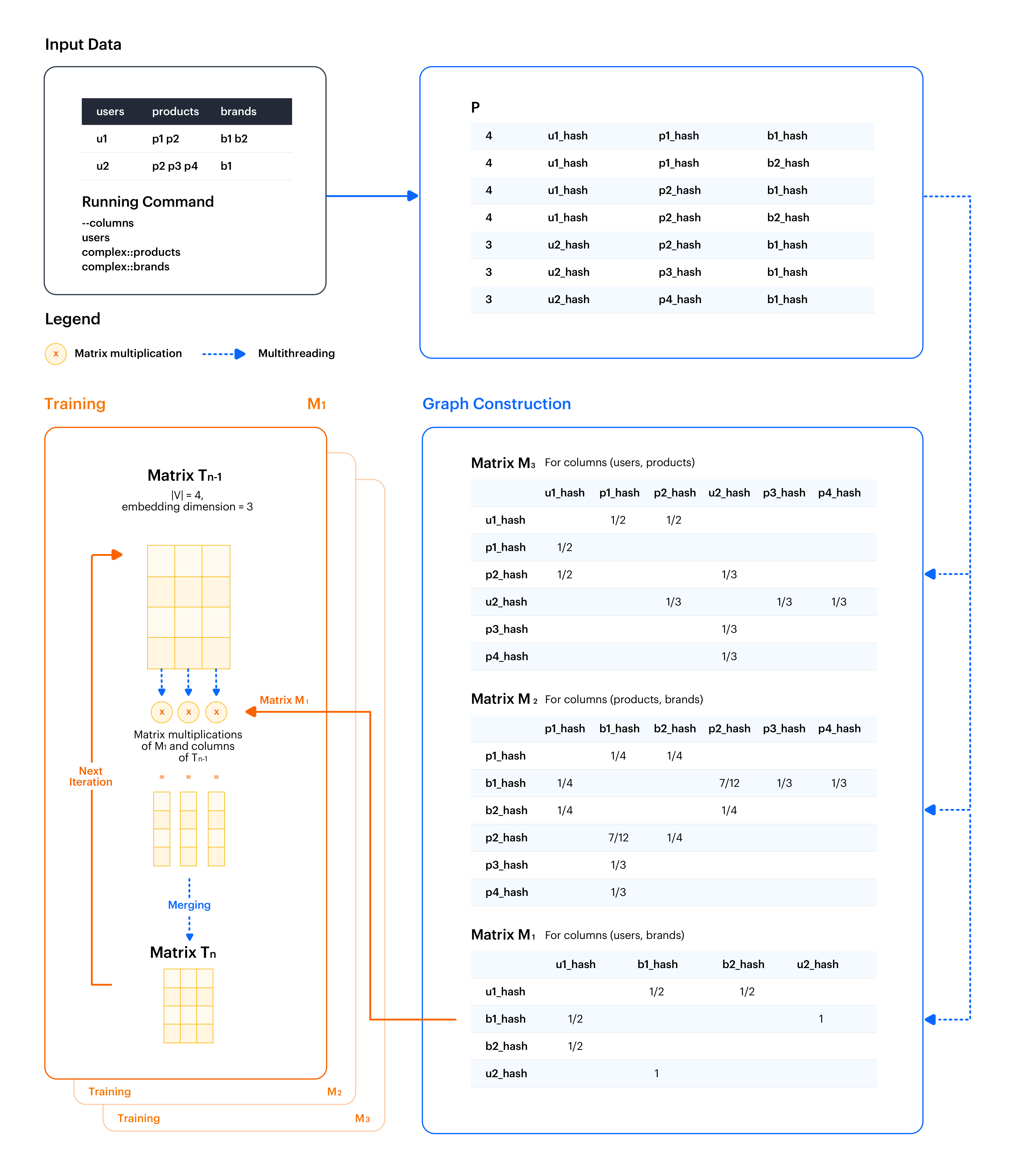}
\caption{Cleora Architecture.}
\label{fig_implementation}
\end{figure*}

Graph construction starts with the creation of a helper matrix $\textbf{P}$ object as a regular 2-D Rust array, which represents the input graph edges  according to the selected expansion method (clique, star, or no expansion). An example involving clique expansion is presented in Figure \ref{fig_implementation} - a Cartesian product (all combinations) of all columns is created. If there are more than 2 relations in the graph, multiple separate $\textbf{M}$ matrices will be created for each relation pair. 

Each entity identifier from the original input file is hashed with \texttt{xxhash}\footnote{\url{https://cyan4973.github.io/xxHash/}} - a fast and efficient hashing method. We hash the identifiers to store them in a unified, small data format.

Subsequently, for each relation pair from matrix $\textbf{P}$ we create a separate matrix $\textbf{M}$ as a  \texttt{SparseMatrix} object (the matrices $\textbf{M}$ will usually hold mostly zeros). Each matrix $\textbf{M}$ object is produced in a separate thread.

\subsection{Training}
In this step training proceeds separately for each matrix $\textbf{M}$, so we will now refer to a single object $\textbf{M}$. The matrix $\textbf{M}$ is multiplied by a freshly initialized 2-D array representing matrix $\mathbf{T}_0$ - this array will represent node embeddings. Multiplication is done against each column of matrix $\mathbf{T}_0$ object in a separate thread. The obtained columns of the new matrix $\mathbf{T}_1$ object are subsequently merged into the full matrix. $\mathbf{T}_1$ matrix is L2-normalized, again in a multithreaded fashion across matrix columns. The appropriate matrix representation lets us accelerate memory access taking advantage of CPU caching.

Finally, depending on the target iteration number, the $\mathbf{T}_1$ matrix object is either returned as program output and printed to file, or fed for next iterations of multiplication against the matrix $\textbf{M}$ object.

\subsection{Memory Consumption}

Memory consumption is linear to the number of nodes and edges. To be precise, during training we need to allocate space for the following:

\begin{itemize}
    \item $|V|$ objects of 40 bytes to store the matrix $\textbf{P}$;
    \item $2 \times |E|$ objects of 24 bytes (in undirected graphs we need to count an edge in both directions) to store the matrix $\textbf{M}$;
    \item $2 \times d \times |V|$ objects of 4 bytes to store the embedding matrix $\textbf{T}$.
\end{itemize}

As such, the total memory complexity is given by $O(|V|(1+2d) + 2|E|)$.
 
 \begin{table*}
\small
\centering
\begin{tabular}{ l|c|c|c|c|c } 
 \hline

 Name & \textbf{Facebook} & \textbf{YouTube} & \textbf{RoadNet} & \textbf{LiveJournal} & \textbf{Twitter} \\\hline
 \#Nodes & 22,470 & 1,134,890 & 1,965,206 & 4,847,571 & 41,652,230 \\ 
 \#Edges & 171,002 & 2,987,624 & 2,766,607 & 68,993,773 & 1,468,365,182 \\ 
 Average Degree & 12 & 5 & 3 & 16 & 36 \\
 Density & $6\times10^{-4}$ & $4.7\times10^{-6}$ & $1.4\times10^{-6}$ & $3.4\times10^{-6}$ & $9.1\times10^{-7}$\\
 \#Classes & 4 & 47 & - & - & - \\
 Directed & No & No & No & Yes & Yes \\
 \hline
\end{tabular}
\caption{Dataset characteristics.}
\label{datasets}
\end{table*}

\begin{table*}
\small
\centering
\begin{tabular}{ l|c|c|c|c|c } 
 \hline
 Algorithm & \textbf{Facebook} & \textbf{YouTube} & \textbf{RoadNet} & \textbf{LiveJournal} & \textbf{Twitter} \\\hline
  \multicolumn{6}{c}{\textbf{Total embedding time}} \\\hline
 Cleora & 00:00:43 h & 00:12:07 h & 00:24:15 h & 01:35:40 h  & 25:34:18 h \\ 
 PBG & 00:04.33 h & 00:54:35 h & 00:37:41 h &  10:38:03 h & -* \\
 Deepwalk & 00:36:51 h & 28:33:52 h & 53:29:13 h & \textit{timeout} & \textit{timeout} \\ 
\hline
 \multicolumn{6}{c}{\textbf{Training time}} \\ \hline
 Cleora & 00:00:25 h & 00:11:46 h & 00:04:14 h & 01:31:42 h & 17:14:01 h \\ 
 PBG & 00:02:03 h & 00:24:15 h & 00:31:11 h & 07:10:00 h & -* \\
 \hline
\end{tabular}
\caption{Calculation times of CPU-based methods: Cleora, PBG and Deepwalk. Total embedding times encompass the whole training procedure, including data loading and preprocessing. Training times encompass the training procedure itself, excluding data loading and preprocessing. * - training crashed due to excessive resource consumption.}
\label{training-time}
\end{table*}

 \begin{table*}
\centering

\begin{tabular}{l|ll|ll|ll|ll|ll}
 \multirow{2}{1mm}{\textbf{Algorithm}} & 
 \multicolumn{2}{c}{\textbf{Facebook}} & 
 \multicolumn{2}{c}{\textbf{YouTube}}  &
 \multicolumn{2}{c}{\textbf{RoadNet}}  & 
 \multicolumn{2}{c}{\textbf{LiveJournal}} &
 \multicolumn{2}{c}{\textbf{Twitter}} 
 \\

  & MRR & HR@10 & MRR & HR@10 & MRR & HR@10 & MRR & HR@10 & MRR & HR@10 \\\hline
  
   \multicolumn{11}{c}{Scalable methods} \\
   \hline

Cleora & 0.0724 & 0.1761 & 0.0471 & 0.0618 & 0.9243 & 0.9429 & 0.6079 & 0.6665 & 0.0355 & 0.076 \\
 PBG \cite{pbg} & 0.0817* & 0.2133* & 0.0321* & 0.0640* & 0.8717* & 0.9106* &  0.5669* & 0.6730*  & -** & -** \\
 GOSH \cite{10.1145/3404397.3404456} &  0.0924* & 0.2319* & 0.0280* & 0.0590* & 0.8756* & 0.8977* & 0.2242* & 0.4012*  & -** & -** \\
\hline
   \multicolumn{11}{c}{Non-scalable methods} \\
   \hline
 Deepwalk \cite{perozzi2014deepwalk} & 0.0803* & 0.1451* & 0.1045* & 0.1805* & 0.9626* & 0.9715* & \textit{timeout} & \textit{timeout} & \textit{timeout} & \textit{timeout} \\ 
 LINE \cite{tang2015line} & 0.0749* & 0.1923* & 0.1064* & 0.1813* & 0.9628* & 0.9833* & 0.5663* & 0.6670* & -** & -**\\
 \hline
\end{tabular}
\caption{Link prediction performance results. * - results with statistically significant differences to Cleora according to the Wilcoxon two-sided paired test (p-value lower than 0.05). ** - training crashed due to excessive resource consumption/unscalable architecture.}
\label{performance-lp}
\end{table*}

\begin{table}
\centering
\begin{tabular}{l|ll|ll}
 \multirow{2}{1mm}{\textbf{Algorithm}} & \multicolumn{2}{l}{\textbf{Facebook}} & \multicolumn{2}{l}{\textbf{YouTube}} \\
  & Micro-F1 & Macro-F1 & Micro-F1 & Macro-F1 \\\hline
  
     \multicolumn{5}{c}{Scalable methods} \\
   \hline
Cleora & 0.9165 & 0.9166 & 0.3859 & 0.3077 \\
 PBG & 0.9258 & 0.9262 & 0.3567* & 0.2459* \\
 GOSH & 0.8312* & 0.8305* & 0.3166* & 0.2245*  \\
 \hline
 \multicolumn{5}{c}{Non-scalable methods} \\
   \hline
 Deepwalk & 0.9349* & 0.9354* & 0.3166* & 0.2245*  \\
 LINE & 0.9442* & 0.9446* & 0.4008* & 0.3338* \\
 \hline
\end{tabular}
\caption{Classification performance results. * - results with statistically significant differences to Cleora according to the Wilcoxon two-sided paired test (p-value lower than 0.05).}
\label{performance-classification}
\end{table}

\begin{table*}
\small
\centering
\begin{tabular}{l|lll|lll}
 \multirow{2}{1mm}{\textbf{Algorithm}} & \multicolumn{3}{c}{\textbf{Facebook}} & \multicolumn{3}{c}{\textbf{YouTube}} \\
  & Micro-F1 & Macro-F1 & MR & Micro-F1 & Macro-F1 & MR \\\hline
Original & 0.9190 & 0.9191 & 457 & 0.3673 & 0.3032 & 3677 \\
1-hop reconstruction & 0.8718 & 0.8586 & 704 & 0.3013 & 0.2046 & 3751 \\
2-hop reconstruction & 0.7856 & 0.7664 & 1552 & 0.2195 & 0.1136 & 3943 \\
 \hline
\end{tabular}
\caption{Quality check of reconstructed nodes in a challenging setting where only 30\% of all nodes are learned directly and 70\% are reconstructed. 1-hop reconstruction nodes are computed from original nodes. 2-hop reconstruction nodes are computed from 1-hop reconstruction nodes.}
\label{performance-reconstruction}
\end{table*}

\begin{figure*}
     \centering
     \begin{subfigure}[b]{0.45\textwidth}
         \centering
         \includegraphics[width=\textwidth]{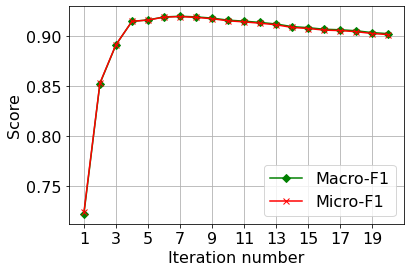}
         \caption{Facebook Dataset - Classification Task.}
         \label{fig:y equals x}
     \end{subfigure}
\hfill
     \begin{subfigure}[b]{0.45\textwidth}
         \centering
         \includegraphics[width=\textwidth]{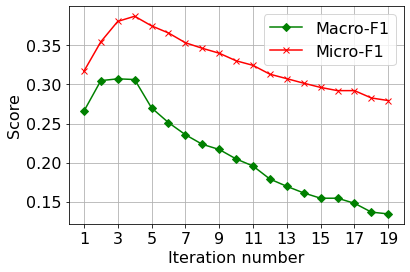}
         \caption{Youtube Dataset - Classification Task}
         \label{fig:three sin x}
     \end{subfigure}
     \hfill
     \begin{subfigure}[b]{0.45\textwidth}
         \centering
         \includegraphics[width=\textwidth]{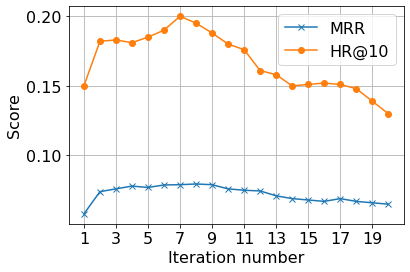}
         \caption{Facebook Dataset - Link Prediction Task.}
         \label{fig:five over x}
     \end{subfigure}
     \hfill
     \begin{subfigure}[b]{0.45\textwidth}
         \centering
         \includegraphics[width=\textwidth]{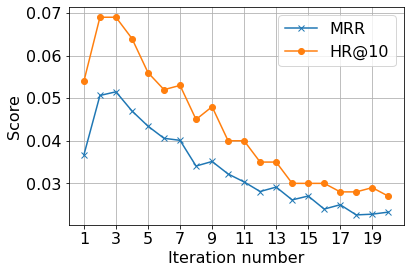}
         \caption{Youtube Dataset - Link Prediction Task.}
         \label{fig:five over x}
     \end{subfigure}
        \caption{The influence of iteration number on embedding quality.}
        \label{fig:iternum}
\end{figure*}

\begin{figure*}
     \centering
     \begin{subfigure}[b]{0.3\textwidth}
         \centering
         \includegraphics[width=\textwidth]{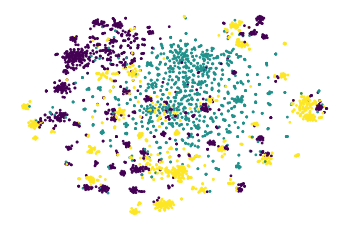}
         \caption{Cleora}
     \end{subfigure}
\hfill
     \begin{subfigure}[b]{0.3\textwidth}
         \centering
         \includegraphics[width=\textwidth]{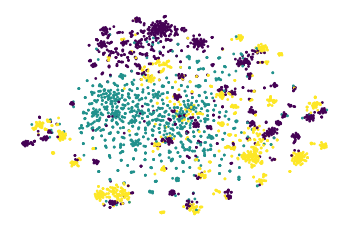}
         \caption{GOSH}
     \end{subfigure}
     \hfill
     \begin{subfigure}[b]{0.3\textwidth}
         \centering
         \includegraphics[width=\textwidth]{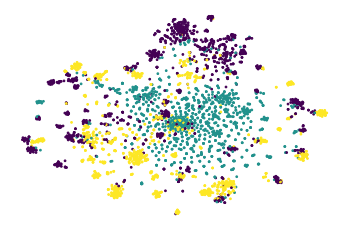}
         \caption{PBG}
     \end{subfigure}

        \caption{t-SNE projections of embeddings learned on Facebook dataset, colored by class assignment. The displayed methods are: a) Cleora b) GOSH c) PBG.}
        \label{fig:t-sne}
\end{figure*}

\section{Experiments}
To evaluate the quality of Cleora embeddings, we study their performance on two popular tasks: link prediction \cite{10.5555/1241540.1241551} and node classification. As competitors we consider recent state-of-the-art methods designed for big data: PBG and GOSH, as well as classic models: LINE and Deepwalk. The details on the competitors' algorithms are described in Section \ref{related-work} (Related Work). For each competitor, we use its original author's implementation. We train each model with the best parameter configurations reported in their original papers or repositories per each dataset. For datasets which do not have a per-model best configuration, we perform an experimental grid search.

We conduct all experiments on two machines: 1) Standard E32s v3 Azure  (32 vCPUs/16 cores) and 256 GB RAM for CPU-based methods, and 2) 128 GB RAM, 14 core (28 HT threads) Intel Core i9-9940X 3.30GHz CPUs and  GeForce RTX 2080 Ti 11GB GPU for GPU-based methods.

\subsection{Datasets}
We use 5 public datasets, summarized in Table \ref{datasets}. We focus on picking the most referential datasets, which are already popular benchmarks. Moreover, we take care to include datasets of various sizes, spanning from medium-sized (Facebook) to massive (Twitter). For each experiment we randomly sample out 80\% of the edges found in each dataset as trainset (both for the embeddings and the proceeding proxy tasks) and the rest serves as validation/test dataset.

\begin{itemize}
    \item \textbf{Facebook} \cite{rozemberczki2019multiscale}. In this graph, the nodes represent official Facebook pages while the links are mutual likes between sites. All nodes belong to one of 4 classes defined by Facebook: politicians, governmental organizations, television shows and companies.
    \item \textbf{Youtube} \cite{mislove-2007-socialnetworks}. This graph represents mutual friendships in the Youtube social networks. Nodes are also characterized by class assignments representing membership in groups.
    \item \textbf{RoadNet} \cite{community-structure}. A road network of California. Intersections and endpoints are represented by nodes and the roads connecting these intersections or road endpoints are represented by edges.
    \item \textbf{LiveJournal} \cite{10.1145/1150402.1150412}. LiveJournal is a free online community with almost 10 million members. The graph represents friendships among members.
    \item \textbf{Twitter} \cite{Kwak10www}. A subset of the Twitter network with directed follower-following social network. This graph is marked by low reciprocity of relations and a strong presence of influential nodes.
\end{itemize}

\subsection{Embedding Computation}
We use the following configurations for each of the models:
\begin{itemize}
    \item Cleora: all embeddings are trained with iteration number $i$ = 4 and embedding dimensionality $d$ = 1024. We treat each row of the adjacency matrix as a hyperedge. Hyperedges are expanded with the clique expansion approach in Facebook and RoadNet datasets, and with star expansion in YouTube and LiveJournal datasets (due to the appearance of very large clusters).
    \item PBG \cite{pbg}: We use the code from original author repository\footnote{\url{https://github.com/facebookresearch/PyTorch-BigGraph/}}. We use the following parameter configurations which we find to give the best results:
    \begin{itemize}
    \item LiveJournal dataset: we reuse the original configuration from the author repository. 
    
    \item Facebook, Roadnet datasets: 
    
        \texttt{dimension=1024},
    
        \texttt{global\_emb=False},
        
        \texttt{num\_epochs=30},
        
        \texttt{lr=0.001},
        
        \texttt{regularization\_coef=1e-3},
        
        \texttt{comparator="dot"}  
        
    \item Youtube dataset: same as above, with
    
    \texttt{num\_epochs=40}
    
    \end{itemize}
    For link prediction we use \texttt{loss\_fn="ranking"} and for classification \texttt{loss\_fn="softmax"} as suggested in the documentation.

    \item LINE \cite{tang2015line}: We use the code from Graphvite\footnote{\url{https://github.com/DeepGraphLearning/graphvite}}, which includes a parallelized GPU-based implementation of the original algorithm. We use the configuration for Youtube dataset from author repository for embedding this dataset. We reuse this configuration for other datasets with epoch number $e$ = 4000. We concatenate the base embeddings with context embeddings as recommended by the authors.
    \item GOSH \cite{10.1145/3404397.3404456}: We use the original author code\footnote{\url{https://github.com/SabanciParallelComputing/GOSH}}. The model is trained with learning rate $lr$ = 0.045 and epochs $e$ = 1000, defined in the original paper as the optimal configuration for our graph sizes (without any graph coarsening for maximum accuracy). 
    \item Deepwalk \cite{perozzi2014deepwalk}: We use the original author code\footnote{\url{https://github.com/phanein/deepwalk}}. We reuse the optimal configuration from author paper: 
    \texttt{-workers 30},
    
    \texttt{--representation-size 128},
    
    \texttt{--number-walks 80},
    
    \texttt{--walk-length 40}

\end{itemize}

We compare the results of scalable methods (Cleora, PBG, GOSH), contrasting them with unscalable methods (Deepwalk, LINE). We deem LINE unscalable as its implementation cannot embed a graph when the total size of the embedding is larger than the total available GPU memory.

Embedding training times are displayed in Table \ref{training-time}. We include only CPU-based methods for fair comparison. PBG is the fastest high-performing CPU-based method we are aware of, and Deepwalk serves as a baseline example of the non-scalable methods. Cleora is shown to be significantly faster than PBG in each case. Training is up to 5 times faster than PBG and over 200 times faster than Deepwalk.

We face technical difficulties with training the embeddings on the Twitter graph. Both GOSH and PBG crash during or just after  loading the data into memory due to exhausting the available hardware resources.
This demonstrates that Cleora can be trained on hardware which fails for other fast methods (note, though, that our hardware configuration is not small but rather standard). Also, technical errors can stem directly from complex model architecture (e.g. heavy and complicated parallelism in PBG), a problem which does not exist in Cleora due to the simplicity of the algorithm. As a consequence, the only model we evaluate on Twitter is Cleora.

\subsection{Performance Check}
\textbf{Link Prediction.}
Link prediction is one of the most common proxy tasks for embedding quality evaluation \cite{liben2007link}. It consists in determining whether two nodes should be connected with an edge. 


 We collect the embeddings for all positive pairs (real edges) and for each such pair we generate a negative pair with random sampling. We compute the Hadamard product for each pair, which is a standard procedure observed to work well for performance comparison \cite{grover2016node2vec, 10.1145/3184558.3191523}. Then, we feed the products to a Logistic Regression classifier which predicts the existence of edges. As the graphs we use are mostly large-scale, we resort to step-wise training with the \texttt{SGDClassifier} module from \texttt{scikit-learn} with a Logistic Regression solver.
 
 For evaluation, we reuse the setting from \cite{pbg}: for each positive  node pair, we pick 10,000 most popular graph nodes, which are then paired with the start node of each positive pair to create negative examples. We rank the pairs, noting the position of the real pair among the 10,001 total examples. In this setting we are able to compute popular ranking-based metrics:
\begin{itemize}
    \item MRR (mean reciprocal rank). This is the average of the reciprocals of the ranks of all positives. The standard formulation of MRR requires that only the single highest rank be summed for each example $i$ (denoted as $rank_i$) from the whole query set $Q$:

        $$ MRR = \frac{1}{|Q|} \sum_{i=1}^{|Q|}{\frac{1}{rank_i}}. $$ 

    As we consider all edge pairs separately, this is true by default in our setting. Higher score is better, the best is 1.0.
    
    \item HitRate@10. This gives the percentage of valid end nodes which rank in top 10 predictions among their negatives. Higher score is better, the best score is 1.0.
\end{itemize}

Due to a large size of graphs, we evaluate on 100,000 sampled examples from the testset. The variances of scores among  samples drawn with various seeds are very low, on the level of $1\mathrm{e}{-8}$ to $1\mathrm{e}{-12}$, so the sampling error will have no visible influence on model ranking.

Results of our evaluation versus the competitors are presented in Table \ref{performance-lp}. The table shows that in spite of its extreme simplicity, the quality of Cleora embeddings is generally on par with the scalable competitors. Cleora performs worse than the competitors on the smallest graph - Facebook, but gets better with increasing graph size. In some cases Cleora reaches results that are better than other fast algorithms aligned for big data - PBG and GOSH. Cleora performs especially well in the MRR metric, which means that it positions the correct node pair high enough in the result list as to avoid punishment with the harmonically declining scores.

Apart from score rankings, another important issue is embedding versatility. This can be problematic in particular for PBG, which defines a number of loss functions linked to the downstream task. Indeed, we note that in order to achieve the top results we observe, it is necessary to train 2 versions of the embeddings: a version for link prediction task with \texttt{loss\_fn="ranking"} and a version for classification task with \texttt{loss\_fn="softmax"}. Whenever training with the classification objective, we observe significant drops of performance in link prediction (e.g. up to 5 times lower MRR and HR for Youtube dataset). The other models, including Cleora, do not require separate task-specific parameters.

We are unable to compare our results on the Twitter graph to competitors, thus presenting the results only for Cleora in Table \ref{performance-lp}. Due to large size of the graph and the computed embedding file (over 500 GB, far exceeding our available memory size), we compute the results on a large connected component of the graph, containing 4,182,829 nodes and 8,197,949 edges.

\textbf{Classification.} We evaluate node classification on two datasets which have node labels: Facebook and YouTube. In Facebook, we use the 4 available classes. In YouTube, the number of classes (representing interest group numbers) is on the scale of thousands. We select 47 most numerous classes for our classification objective (following \cite{tang2015line}). As performance measures we use micro-F1 and macro-F1 scores from \texttt{scikit-learn} library.

The results are displayed in Table \ref{performance-classification}. The findings confirm the results from link prediction. Cleora is again ranking high, this time usually beating the scalable competitors, and getting close to non-scalable ones.

Figure \ref{fig:t-sne} shows Cleora, GOSH and PBG vectors learned on Facebook and projected to 2-D space using t-SNE \cite{vanDerMaaten2008}. The vectors are colored by their class assignments. It can be observed that Cleora learns an essentially similar data abstraction as the contrastive methods. Additionally, a number of outliers appear in Cleora's visualization, which are nodes from small, strongly interconnected clusters. Such nodes' embeddings get extremely similar to each other and at the same time distant from the rest of the embeddings. As such, Cleora clearly separates strongly connected components of the graph.

\textbf{Node Reconstruction (Inductivity).} We also evaluate the quality of embeddings of new nodes which are added to the graph after the computation of the full graph embedding table. To this end, we split the Facebook and Youtube datasets into a 'seen node' set comprised of 30\% of each dataset, and 'new node' set comprised of the remaining 70\% of each dataset. This setting is challenging as the vast majority of nodes will need to be reconstructed. At the same time, it gives us enough data to study embedding quality of two kinds of reconstruction: \textbf{1-hop reconstruction}, where a new node embedding is created based on 'seen node' embeddings, which have been learned directly from interaction data, and \textbf{2-hop reconstruction}, where a new node embedding is created based on reconstructed node embeddings. 2-hop reconstruction is a much harder task, as errors from previous reconstruction step will be accumulated.

Table \ref{performance-reconstruction} shows the results of the reconstruction quality check. We use the previous measures of micro-F1 and macro-F1 in the Node Classification task, also taking into account the MR (Mean Rank) measure, which is a simple average of obtained ranks in the Link Prediction task. The measure is a simplified abstraction of the MRR and HR metrics. 1-hop and 2-hop reconstruction results are computed on the reconstructed embeddings only.

We observe that the task of classification on Facebook dataset notes only a 5\% micro-F1 drop in 1-hop node generation, and a 14.5\% micro-F1 drop in 2-hop node generation. Youtube faced a stronger drop of 18\% and 40\%, respectively. A reversed tendency is observed for link prediction: the MR drops are severe for Facebook (MR falling by a half when reconstructed), while being very mild for Youtube (absolute difference in MR being only 266). This suggests that quality of reconstruction is heavily dependent on dataset. Even with 2-hop reconstruction, it is possible to obtain embeddings of  high relative quality.

\begin{table*}
    \centering

\begin{tabular}{l|lll|lll} 
 & \multicolumn{3}{c}{\textbf{butterfly}}
 & \multicolumn{3}{c}{\textbf{elephant}} \\

 & w2v & Cleora $i$ = 1 & Cleora $i$ = 4
 & w2v & Cleora $i$ = 1 & Cleora $i$ = 4
 \\\hline
 
 1. & feather & madama & cover & giraffe & dumbo & elephants  \\                        
 2. & shark & wildflower & pictured & orca & elephas & mammoths \\                      
 3. & unicorn & state & strange & panda & elephantiasis & hunting \\                      
 4. & eyed & norsemen & photograph &  ox & loxodonta & predator   \\
 5. & pear & gopal & nice & albino & shrews & covered \\ 
 \hline
 & \multicolumn{3}{c}{\textbf{king}} 
 & \multicolumn{3}{c}{\textbf{nationalism}} \\
 & w2v & Cleora $i$ = 1 & Cleora $i$ = 4 
 & w2v & Cleora $i$ = 1 & Cleora $i$ = 4 \\\hline
 
 1. & prince & coretta & iii & imperialism & eurocentrism & colonialism \\ 
 2. & queen & fahd & conqueror & nazism & chauvinism & political \\                    
 3. & throne & aleksandar & succeeded & ideology & simmering & domination\\ 
 4. & antiochus & anshan & queen & fascism & republicanism & imperialism\\          
 5. & kings & sihamoni & lombard & conservatism & arabization & establishment\\

\end{tabular} 
\caption{Results of the homophily vs. structural equivalence experiment. We contrast Cleora with Word2Vec \cite{Word2Vec} showing varied levels of homophily and structural equivalence depending on iteration number.}
\label{tab-struct-vs-functional}
\end{table*}

\section{Analysis}

\subsection{Optimal iteration number}
The iteration number defines the breadth of neighborhood on which a single node is averaged: iteration number $i$ means that nodes with similar $i$-hop neighborhoods will have similar representations. We show the influence of $i$ on classification performance in Figure \ref{fig:iternum}.

The iteration number is related to the concept of average path length from the area of graph topology. The average path length is defined as the average number of steps along the shortest paths for all possible pairs of network nodes. If the iteration number reaches the average path length, an average node will likely have access to all other nodes. Thus, iteration number slightly exceeding the average path length can be deemed optimal. For example, the average path length for the Facebook graph equals 5.18 and the best $i$ is found to be 5-7 according to Figure \ref{fig:iternum}. In practice however the computation of average path length is significantly slower than the computation of Cleora embeddings. An optimal solution to determining the iteration number is to verify iteration number empirically on a downstream task. 

Too large iteration number will make all embeddings gradually more similar to each other, eventually collapsing to an exact same representation. This behavior might be rather slow or abrupt after passing the optimal point depending on the dataset, as evidenced in Figure \ref{fig:iternum}.

\subsection{Homophily and Structural Equivalence}
The idea of node similarity can be considered under various formulations. \cite{grover2016node2vec} propose to focus on two such aspects: homophily \cite{doi:10.1146/annurev.soc.27.1.415} and structural equivalence. Under the homophily criterion, node embeddings should be similar when the nodes are highly interconnected. Structural equivalence, on the other hand, stresses that similarity of node embeddings should be based on their role within a graph (e.g. two nodes which are centers of large clusters should be similar, or two leaf nodes should be similar). Under this definition, Cleora leans strongly towards the homophily criterion. An extreme example of this behavior would be the embedding of a leaf node. Leaf nodes have only one neighbor, so their embedding is always equal to the embedding of their single neighbor from the previous iteration. As such, two leaf nodes will have very different embeddings if their geodesic distance is higher than the iteration number. With rising iteration number, more structural knowledge flows into the embeddings.

We exemplify this property contrasting Cleora with Word2Vec \cite{Word2Vec} in Table \ref{tab-struct-vs-functional}. Both models are trained on the \texttt{enwik8} dataset\footnote{\url{https://deepai.org/dataset/enwik8}}. Cleora is trained with clique expansion approach on token sequences with window sizes of 5,7,9. Representations computed with each window size were concatenated. 

We query each set of embeddings with selected nouns to find their nearest neighbors. Word2Vec embeddings exhibit strong structural equivalence, usually keeping to the part of speech (PoS) of the query. PoS is an important structural characteristic of a word, defining its role within a sentence. On the other hand, Cleora $i$ = 4 usually returns similar nouns but sometimes also related words of other PoS, e.g \texttt{elephant} is related to \texttt{hunting}. Cleora $i$ = 1 is an extreme case of homophily in close neighborhoods, returning words which appear in one phrase, e.g. \texttt{madama butterfly}, \texttt{elephant dumbo}.

\subsection{Complement and Substitute Prediction}
Node similarity problem can be cast as complement versus substitute identification \cite{mcauley2015inferring}, which is especially interesting in e-commerce applications. Substitutes are defined as products that can be purchased instead of each other, while complements are products that can be purchased in addition to each other (they often appear together in the same basket). As such, complements are expected to be similar if their 1-hop neighborhoods are similar, boiling down to neighborhood prediction. Substitutes should be close in terms of a broader notion of semantic similarity of nodes.

\begin{table*}[h!]

    \centering
    \scalebox{0.9}{
    \begin{tabular}{l|ll} 
    
 & \multicolumn{2}{c}{\textbf{SOUP RAMEN NOODLES/RAMEN CUPS 3 OZ}} \\
 & 1 iteration (Complement) & 4 iterations (Substitute) \\\hline
1. & AUTOMOTIVE PRODUCTS 4 CT & SOUP RAMEN NOODLES/RAMEN CUPS 3 OZ\\
2. & PROCESSED DIPS 15.5 OZ & SOUP RAMEN NOODLES/RAMEN CUPS 3 OZ \\
3. & SOUP RAMEN NOODLES/RAMEN CUPS 3 OZ & SOUP RAMEN NOODLES/RAMEN CUPS 3 OZ \\
4. & J-HOOKS JHOOK - HOUSEWARE & SOUP RAMEN NOODLES/RAMEN CUPS 3 OZ \\  
5. & PACKAGED CANDY BAGS-CHOCOLATE 11 OZ & SOUP RAMEN NOODLES/RAMEN CUPS 3 OZ \\

& \multicolumn{2}{c}{\textbf{BAKED BREAD/BUNS/ROLLS MAINSTREAM WHITE BREAD 20 OZ}} \\
& 1 iteration (Complement) & 4 iterations (Substitute) \\\hline
1. & BAKED BREAD/BUNS/ROLLS DINNER ROLLS 11 OZ & SMOKED MEATS MARINATED  \\
2. & CHIPS\&SNACKS MISC 3.5 OZ & PICKLE/RELISH/PKLD VEG PICKLES \\
3. & SPRING/SUMMER SEASONAL SALLY HANSEN & PNT BTR/JELLY/JAMS JELLY \\  
4. & DRY NOODLES/PASTA SPAGHETTI DRY 16 OZ & COLD CEREAL KIDS CEREAL \\
5. & BEERS/ALES BEERALEMALT LIQUORS 40 OZ & BREAKFAST SAUSAGE/SANDWICHES PATTIES \\
& \multicolumn{2}{c}{\textbf{BREAD BREAD:ITALIAN/FRENCH }} \\
& 1 iteration (Complement) & 4 iterations (Substitute) \\\hline
1. & LUNCHMEAT PEPPERONI/SALAMI 3 OZ &  REFRIGERATED DOUGH PRODUCTS  ROLLS \\
2. & CANDY BAGS-NON CHOCOLATE 4.25 OZ & SEAFOOD - FROZEN SEAFOOD-FRZ-RW-ALL \\
3. & GREETING CARDS/WRAP/PARTY SPLY PARTY & BAKED SWEET GOODS SNACK CAKE - PACK 5.7 OZ \\
4. & VALENTINE VALENTINE GIFTWARE/DECOR 5 CT & PIES PIES: PUMPKIN/CUSTARD   \\  
5. & CANDY - CHECKLANE CANDY BARS (SINGLES) & LUNCHMEAT HAM 9 OZ 

\end{tabular} 
}
\caption{Examples of complement vs. substitute prediction on shopping baskets from Dunnhumby dataset. We show that low iteration numbers produce embeddings which reflect the complement relation. On the other hand, high iteration numbers produce embeddings which reflect the substitute relation.}
\label{tab:dunnhumby}
\end{table*}

Thanks to the influence of the iteration number, Cleora can represent either the complementary or the substitute relation. To exemplify this property, we preprocess the "Complete Journey" Dunnhumby dataset\footnote{\url{https://www.dunnhumby.com/source-files/}}. The dataset includes transactions of 2,500 households within all categories in the store, gathered over 2 years. We embed the product baskets with Cleora, expanding each basket with the clique expansion approach. We use two configurations: iteration number $i$ = 1 and $i$ = 4, and compute the nearest neighbors for each product using cosine similarity.

Extracts from our results are displayed in Table \ref{tab:dunnhumby}. With $i$ = 1 the closest embeddings mirror the complement relation. For example, instant Ramen soups seem to be often bought together with automotive parts, dips and sweets which suggests that they might be bought by drivers stopping by. Regarding bread, there seems to be a marked difference in the perception of regular white bread and French bread. White bread is associated with everyday meals, while French bread is bought together with greeting cards, candy and gifts. As such, it is much more strongly connected to special occasions such as celebration of Valentine's Day and parties. In the case of $i$ = 4, products are more closely aligned with similar products (possible substitutes): ramen soups with other ramen soups of various kinds (only the product IDs differ here). Breads are close to cereal, pies, sandwiches and other bakery products but also meats.

Note that nearest neighbor search between $i$ = 1 embeddings does not achieve identical goals as algorithms of association rule mining \cite{agrawal1993mining, agarwal1994fast, grahne2003efficiently}. Due to the averaging operation over neighbors, embeddings of items appearing in small baskets will be much more similar to each other than if they appeared in bigger baskets. Thus, nearest neigbor search with Cleora strongly promotes smaller, more selective clusters.

\section{Summary}

We have presented Cleora - a simple, purely unsupervised embedding algorithm which learns representations analogous to contrastive methods. Cleora is much faster than other CPU-based methods. Moreover, it has useful extra properties, such as node embedding inductivity and the ability to compute partial embeddings on chunked graph, which can be subsequently merged. The iteration number allows to obtain various functionalities, for example complement or substitute prediction. We open-source Cleora to the community in order to aid reproducibility and allow a wide use of our method, including commercial use.


\begin{thebibliography}{00}

\bibitem{agarwal1994fast}
Rakesh Agarwal, Ramakrishnan Srikant, et~al.
\newblock Fast algorithms for mining association rules.
\newblock In {\em Proc. of the 20th VLDB Conference}, pages 487--499, 1994.

\bibitem{agrawal1993mining}
Rakesh Agrawal, Tomasz Imielinski, and Arun Swami.
\newblock Mining associations between sets of items in large databases.
\newblock In {\em Proceedings of the ACM SIGMOD International Conference on
  Management of Data}, pages 207--216, 1993.

\bibitem{10.1145/3404397.3404456}
Taha~Atahan Akyildiz, Amro~Alabsi Aljundi, and Kamer Kaya.
\newblock Gosh: Embedding big graphs on small hardware.
\newblock In {\em 49th International Conference on Parallel Processing - ICPP},
  ICPP '20, New York, NY, USA, 2020. Association for Computing Machinery.

\bibitem{aletras2018predicting}
Nikolaos Aletras and Benjamin~Paul Chamberlain.
\newblock Predicting twitter user socioeconomic attributes with network and
  language information.
\newblock In {\em Proceedings of the 29th on Hypertext and Social Media}, pages
  20--24. 2018.

\bibitem{asatani2018detecting}
Kimitaka Asatani, Junichiro Mori, Masanao Ochi, and Ichiro Sakata.
\newblock Detecting trends in academic research from a citation network using
  network representation learning.
\newblock {\em PloS one}, 13(5):e0197260, 2018.

\bibitem{10.1145/1150402.1150412}
Lars Backstrom, Dan Huttenlocher, Jon Kleinberg, and Xiangyang Lan.
\newblock Group formation in large social networks: Membership, growth, and
  evolution.
\newblock In {\em Proceedings of the 12th ACM SIGKDD International Conference
  on Knowledge Discovery and Data Mining}, KDD '06, page 44–54, New York, NY,
  USA, 2006. Association for Computing Machinery.

\bibitem{NIPS2001_1961}
Mikhail Belkin and Partha Niyogi.
\newblock Laplacian eigenmaps and spectral techniques for embedding and
  clustering.
\newblock In T.~G. Dietterich, S.~Becker, and Z.~Ghahramani, editors, {\em
  Advances in Neural Information Processing Systems 14}, pages 585--591. MIT
  Press, 2002.

\bibitem{bordes2013translating}
Antoine Bordes, Nicolas Usunier, Alberto Garcia-Duran, Jason Weston, and Oksana
  Yakhnenko.
\newblock Translating embeddings for modeling multi-relational data.
\newblock {\em Advances in neural information processing systems},
  26:2787--2795, 2013.

\bibitem{10.1145/2806416.2806512}
Shaosheng Cao, Wei Lu, and Qiongkai Xu.
\newblock Grarep: Learning graph representations with global structural
  information.
\newblock In {\em Proceedings of the 24th ACM International on Conference on
  Information and Knowledge Management}, CIKM '15, page 891–900, New York,
  NY, USA, 2015. Association for Computing Machinery.

\bibitem{harp}
Haochen Chen, Bryan Perozzi, Yifan Hu, and Steven Skiena.
\newblock Harp: Hierarchical representation learning for networks.
\newblock In {\em Proceedings of the Thirty-Second AAAI Conference on
  Artificial Intelligence}. AAAI Press, 2018.

\bibitem{cochez2017biased}
Michael Cochez, Petar Ristoski, Simone~Paolo Ponzetto, and Heiko Paulheim.
\newblock Biased graph walks for rdf graph embeddings.
\newblock In {\em Proceedings of the 7th International Conference on Web
  Intelligence, Mining and Semantics}, pages 1--12, 2017.

\bibitem{cox2000multidimensional}
Trevor~F. Cox and M.A.A. Cox.
\newblock {\em Multidimensional Scaling, Second Edition}.
\newblock Chapman and Hall/CRC, 2 edition, 2000.

\bibitem{10.1145/3184558.3191523}
Ayushi Dalmia, Ganesh J, and Manish Gupta.
\newblock Towards interpretation of node embeddings.
\newblock In {\em Companion Proceedings of the The Web Conference 2018}, WWW
  '18, page 945–952, Republic and Canton of Geneva, CHE, 2018. International
  World Wide Web Conferences Steering Committee.

\bibitem{grahne2003efficiently}
G{\"o}sta Grahne and Jianfei Zhu.
\newblock Efficiently using prefix-trees in mining frequent itemsets.
\newblock In {\em FIMI}, volume~90, page~65, 2003.

\bibitem{grover2016node2vec}
Aditya Grover and Jure Leskovec.
\newblock node2vec: Scalable feature learning for networks.
\newblock In {\em Proceedings of the 22nd ACM SIGKDD international conference
  on Knowledge discovery and data mining}, pages 855--864, 2016.

\bibitem{NIPS2017_6703}
Will Hamilton, Zhitao Ying, and Jure Leskovec.
\newblock Inductive representation learning on large graphs.
\newblock In I.~Guyon, U.~V. Luxburg, S.~Bengio, H.~Wallach, R.~Fergus,
  S.~Vishwanathan, and R.~Garnett, editors, {\em Advances in Neural Information
  Processing Systems 30}, pages 1024--1034. Curran Associates, Inc., 2017.

\bibitem{ingraham2019generative}
John Ingraham, Vikas Garg, Regina Barzilay, and Tommi Jaakkola.
\newblock Generative models for graph-based protein design.
\newblock In {\em Advances in Neural Information Processing Systems}, pages
  15820--15831, 2019.

\bibitem{Jolliffe1986}
I.~T. Jolliffe.
\newblock {\em Principal Component Analysis and Factor Analysis}, pages
  115--128.
\newblock Springer New York, New York, NY, 1986.

\bibitem{Kipf:2016tc}
Thomas~N. Kipf and Max Welling.
\newblock {Semi-Supervised Classification with Graph Convolutional Networks}.
\newblock In {\em Proceedings of the 5th International Conference on Learning
  Representations}, ICLR '17, 2017.

\bibitem{Kwak10www}
Haewoon Kwak, Changhyun Lee, Hosung Park, and Sue Moon.
\newblock {W}hat is {T}witter, a social network or a news media?
\newblock In {\em WWW '10: Proceedings of the 19th international conference on
  World wide web}, pages 591--600, New York, NY, USA, 2010. ACM.

\bibitem{pbg}
Adam Lerer, Ledell Wu, Jiajun Shen, Timothee Lacroix, Luca Wehrstedt, Abhijit
  Bose, and Alex Peysakhovich.
\newblock {PyTorch-BigGraph: A Large-scale Graph Embedding System}.
\newblock In {\em Proceedings of the 2nd SysML Conference}, Palo Alto, CA, USA,
  2019.

\bibitem{snapnets}
Jure Leskovec and Andrej Krevl.
\newblock {SNAP Datasets}: {Stanford} large network dataset collection.
\newblock \url{http://snap.stanford.edu/data}, 2014.

\bibitem{community-structure}
Jure Leskovec, Kevin Lang, Anirban Dasgupta, and Michael Mahoney.
\newblock Community structure in large networks: Natural cluster sizes and the
  absence of large well-defined clusters.
\newblock {\em Internet Mathematics}, 6, 11 2008.

\bibitem{10.5555/1241540.1241551}
David Liben-Nowell and Jon Kleinberg.
\newblock The link-prediction problem for social networks.
\newblock {\em J. Am. Soc. Inf. Sci. Technol.}, 58(7):1019–1031, May 2007.

\bibitem{liben2007link}
David Liben-Nowell and Jon Kleinberg.
\newblock The link-prediction problem for social networks.
\newblock {\em Journal of the American society for information science and
  technology}, 58(7):1019--1031, 2007.

\bibitem{mcauley2015inferring}
Julian McAuley, Rahul Pandey, and Jure Leskovec.
\newblock Inferring networks of substitutable and complementary products.
\newblock In {\em Proceedings of the 21th ACM SIGKDD international conference
  on knowledge discovery and data mining}, pages 785--794, 2015.

\bibitem{doi:10.1146/annurev.soc.27.1.415}
Miller McPherson, Lynn Smith-Lovin, and James~M Cook.
\newblock Birds of a feather: Homophily in social networks.
\newblock {\em Annual Review of Sociology}, 27(1):415--444, 2001.

\bibitem{Word2Vec}
Tomas Mikolov, Ilya Sutskever, Kai Chen, Greg~S Corrado, and Jeff Dean.
\newblock Distributed representations of words and phrases and their
  compositionality.
\newblock In {\em Advances in Neural Information Processing Systems}, pages
  3111--3119, 2013.

\bibitem{mislove-2007-socialnetworks}
Alan Mislove, Massimiliano Marcon, Krishna~P. Gummadi, Peter Druschel, and
  Bobby Bhattacharjee.
\newblock {Measurement and Analysis of Online Social Networks}.
\newblock In {\em Proceedings of the 5th ACM/Usenix Internet Measurement
  Conference (IMC'07)}, San Diego, CA, October 2007.

\bibitem{10.1093/bioinformatics/btz600}
Sameh~K Mohamed, Vít Nováček, and Aayah Nounu.
\newblock {Discovering protein drug targets using knowledge graph embeddings}.
\newblock {\em Bioinformatics}, 36(2):603--610, 08 2019.

\bibitem{10.5555/3104482.3104584}
Maximilian Nickel, Volker Tresp, and Hans-Peter Kriegel.
\newblock A three-way model for collective learning on multi-relational data.
\newblock In {\em Proceedings of the 28th International Conference on
  International Conference on Machine Learning}, ICML'11, page 809–816,
  Madison, WI, USA, 2011. Omnipress.

\bibitem{perozzi2014deepwalk}
Bryan Perozzi, Rami Al-Rfou, and Steven Skiena.
\newblock Deepwalk: Online learning of social representations.
\newblock In {\em Proceedings of the 20th ACM SIGKDD international conference
  on Knowledge discovery and data mining}, pages 701--710, 2014.

\bibitem{10.1145/3110025.3110086}
Bryan Perozzi, Vivek Kulkarni, Haochen Chen, and Steven Skiena.
\newblock Don't walk, skip! online learning of multi-scale network embeddings.
\newblock ASONAM '17, page 258–265, New York, NY, USA, 2017. Association for
  Computing Machinery.

\bibitem{pornprasit2020convcn}
Chanathip Pornprasit, Xin Liu, Natthawut Kertkeidkachorn, Kyoung-Sook Kim,
  Thanapon Noraset, and Suppawong Tuarob.
\newblock Convcn: A cnn-based citation network embedding algorithm towards
  citation recommendation.
\newblock In {\em Proceedings of the ACM/IEEE Joint Conference on Digital
  Libraries in 2020}, pages 433--436, 2020.

\bibitem{ristoski2016rdf2vec}
Petar Ristoski and Heiko Paulheim.
\newblock Rdf2vec: Rdf graph embeddings for data mining.
\newblock In {\em International Semantic Web Conference}, pages 498--514.
  Springer, 2016.

\bibitem{rozemberczki2019multiscale}
Benedek Rozemberczki, Carl Allen, and Rik Sarkar.
\newblock Multi-scale attributed node embedding, 2019.

\bibitem{Schumacher2020TheEO}
T.~Schumacher, Hinrikus Wolf, Martin Ritzert, Florian Lemmerich, J.~Bachmann,
  Florian Frantzen, Max Klabunde, M.~Grohe, and M.~Strohmaier.
\newblock The effects of randomness on the stability of node embeddings.
\newblock {\em ArXiv}, abs/2005.10039, 2020.

\bibitem{tang2015line}
Jian Tang, Meng Qu, Mingzhe Wang, Ming Zhang, Jun Yan, and Qiaozhu Mei.
\newblock Line: Large-scale information network embedding.
\newblock In {\em WWW}. ACM, 2015.

\bibitem{10.1145/2736277.2741093}
Jian Tang, Meng Qu, Mingzhe Wang, Ming Zhang, Jun Yan, and Qiaozhu Mei.
\newblock Line: Large-scale information network embedding.
\newblock In {\em Proceedings of the 24th International Conference on World
  Wide Web}, WWW '15, page 1067–1077, Republic and Canton of Geneva, CHE,
  2015. International World Wide Web Conferences Steering Committee.

\bibitem{tenenbaum_global_2000}
Joshua~B. Tenenbaum, Vin de~Silva, and John~C. Langford.
\newblock A global geometric framework for nonlinear dimensionality reduction.
\newblock {\em Science}, 290(5500):2319, 2000.

\bibitem{trouillon2016complex}
Th{\'e}o Trouillon, Johannes Welbl, Sebastian Riedel, {\'E}ric Gaussier, and
  Guillaume Bouchard.
\newblock Complex embeddings for simple link prediction.
\newblock International Conference on Machine Learning (ICML), 2016.

\bibitem{verse}
Anton Tsitsulin, Davide Mottin, Panagiotis Karras, and Emmanuel Müller.
\newblock Verse: Versatile graph embeddings from similarity measures.
\newblock 03 2018.

\bibitem{wu2020learning}
Ning Wu, Xin~Wayne Zhao, Jingyuan Wang, and Dayan Pan.
\newblock Learning effective road network representation with hierarchical
  graph neural networks.
\newblock In {\em Proceedings of the 26th ACM SIGKDD International Conference
  on Knowledge Discovery \& Data Mining}, pages 6--14, 2020.

\bibitem{yang2014embedding}
Bishan Yang, Wen-tau Yih, Xiaodong He, Jianfeng Gao, and Li~Deng.
\newblock Embedding entities and relations for learning and inference in
  knowledge bases.
\newblock {\em arXiv preprint arXiv:1412.6575}, 2014.

\bibitem{yao2019accurately}
Heng Yao, Yunjia Shi, Jihong Guan, and Shuigeng Zhou.
\newblock Accurately detecting protein complexes by graph embedding and
  combining functions with interactions.
\newblock {\em IEEE/ACM Transactions on Computational Biology and
  Bioinformatics}, 17(3):777--787, 2019.

\bibitem{yue2020graph}
Xiang Yue, Zhen Wang, Jingong Huang, Srinivasan Parthasarathy, Soheil
  Moosavinasab, Yungui Huang, Simon~M Lin, Wen Zhang, Ping Zhang, and Huan Sun.
\newblock Graph embedding on biomedical networks: methods, applications and
  evaluations.
\newblock {\em Bioinformatics}, 36(4):1241--1251, 2020.

\bibitem{zhang2018cosine}
Yuan Zhang, Tianshu Lyu, and Yan Zhang.
\newblock Cosine: Community-preserving social network embedding from
  information diffusion cascades.
\newblock In {\em Thirty-second AAAI conference on artificial intelligence},
  2018.

\bibitem{zheng2020gman}
Chuanpan Zheng, Xiaoliang Fan, Cheng Wang, and Jianzhong Qi.
\newblock Gman: A graph multi-attention network for traffic prediction.
\newblock In {\em Proceedings of the AAAI Conference on Artificial
  Intelligence}, volume~34, pages 1234--1241, 2020.

\bibitem{Zhu2019GraphViteAH}
Zhaocheng Zhu, Shizhen Xu, Meng Qu, and Jian Tang.
\newblock Graphvite: A high-performance cpu-gpu hybrid system for node
  embedding.
\newblock {\em The World Wide Web Conference}, 2019.

\bibitem{vanDerMaaten2008}
Laurens van~der Maaten and Geoffrey Hinton.
\newblock Visualizing data using {t-SNE}.
\newblock {\em Journal of Machine Learning Research}, 9:2579--2605, 2008.


\end{thebibliography}
\end{document}